\def\year{2022}\relax
\documentclass[letterpaper]{article} 
\usepackage{aaai22}  
\usepackage{times}  
\usepackage{helvet}  
\usepackage{courier}  
\usepackage[hyphens]{url}  
\usepackage{graphicx} 
\usepackage{natbib}  
\usepackage{caption} 
\DeclareCaptionStyle{ruled}{labelfont=normalfont,labelsep=colon,strut=off} 
\usepackage{algorithm}
\usepackage{algorithmic}

\usepackage{newfloat}
\usepackage{listings}
\floatstyle{ruled}
\newfloat{listing}{tb}{lst}{}
\floatname{listing}{Listing}
 \usepackage{amsmath}
 \usepackage{amssymb}
 \usepackage{amsthm}
 \usepackage{enumerate}
 \usepackage{booktabs}
 \usepackage[switch]{lineno}
 \usepackage{xspace}

\usepackage{ifthen}
\newboolean{proofs}
\setboolean{proofs}{true} 
\newcommand{\Proof}[1]{\ifthenelse{\boolean{proofs}}{\begin{proof}\color{green!50!black} #1 \end{proof}}{}}
\newcommand{\RProof}[1]{\ifthenelse{\boolean{proofs}}{\begin{proof}[\textcolor{magenta}{Proof}]\color{magenta} #1 \end{proof}}{}}

\newcommand{\Omit}[1]{}

\newcommand{\tup}[1]{\langle #1 \rangle}

\newcommand{\citeay}[1]{\citeauthor{#1} (\citeyear{#1})}

\newcommand{\citedate}[1]{(\citeyear{#1})}

\newtheorem{def-thm}[definition]{Definition and Theorem}
\newtheorem{thm-def}[definition]{Theorem and Definition}

\newtheorem*{definition*}{Definition}
\newtheorem*{def-thm*}{Definition and Theorem}
\newtheorem*{thm-def*}{Theorem and Definition}
\newtheorem*{theorem*}{Theorem}
\newtheorem*{lemma*}{Lemma}
\newtheorem*{proposition*}{Proposition}
\newtheorem*{corollary*}{Corollary}

\newenvironment{example-no-eob}{\noindent\textbf{Example.}\,}{}

\newcommand{\Q}{\mathcal{Q}}

\newcommand{\pplus}{\hspace{-.05em}\raisebox{.15ex}{\footnotesize$\uparrow$}}
\newcommand{\mminus}{\hspace{-.05em}\raisebox{.15ex}{\footnotesize$\downarrow$}}

\newcommand{\EQ}[1]{#1{\,=\,}0}
\newcommand{\GT}[1]{#1{\,>\,}0}

\newcommand{\DEC}[1]{#1\mminus}

\newcommand{\UNK}[1]{#1?}

\newcommand{\prule}[2]{\{ #1 \} \mapsto \{ #2 \}}

 \newcommand{\bi}{\begin{itemize}}
 \newcommand{\ei}{\end{itemize}}
 \newcommand{\bii}{\begin{itemize}}
 \newcommand{\eii}{\end{itemize}}

\title{Target Languages (vs. Inductive Biases) for Learning to Act and Plan}

\author{Hector Geffner}

\affiliations{
Universitat Pompeu Fabra, Barcelona, Spain  \\
Instituci\'o Catalana de Recerca i Estudis Avan\c{c}ats (ICREA), Barcelona, Spain\\
Link\"oping University, Link\"oping, Sweden\\
hector.geffner@upf.edu}

\begin{document}


\maketitle

\begin{abstract}
  Recent  breakthroughs in AI have shown the remarkable
  power of deep learning and deep reinforcement learning. These developments, however, have been  tied to specific tasks, 
  and progress  in out-of-distribution generalization  has been  limited.   While it is
  assumed   that these  limitations can be overcome by  incorporating suitable inductive biases,
  the notion of inductive biases itself is often  left vague and does not provide meaningful guidance.
  In the paper, I   articulate a different learning approach  where  representations
  do not emerge from  biases in a neural architecture  but are  learned over a given  target language with  a known  semantics.
   The basic ideas are  implicit   in mainstream AI  where representations have been encoded 
   in languages ranging from fragments of first-order logic to probabilistic structural causal models. 
   The challenge  is  to learn from data, the representations  that have traditionally been crafted by hand.
   Generalization is  then a  result of the semantics  of the language.
  The goals of this paper  are  to  make these  ideas explicit, to  place them   in a broader  context
  where the design  of the target language is crucial,   and  to illustrate them    in the context of learning to  act  and plan.
   For this, after a general discussion, I consider  learning representations of actions,   general policies, and subgoals
   (``intrinsic rewards''). 
  In   these cases,  learning is formulated  as a combinatorial  problem  but nothing   prevents the use of deep learning
  techniques instead. Indeed, learning representations over languages with a known semantics   provides an account of \emph{what}
  is to be learned, while learning representations with neural nets   provides  a complementary account   of \emph{how} representations can be learned.
  The challenge and the opportunity  is to bring the two  together.
\end{abstract}

\section{Introduction}

\Omit{
\begin{quotation}
\emph{If one wants a machine to be able to discover an abstraction, it seems most likely that the machine
must be able to represent this abstraction in some relatively simple way.}  John McCarthy, 1958.

\emph{Systematic generalization is hypothesized to arise from an efficient factorization of knowledge
into recomposable pieces corresponding to reusable factors \ldots This is related yet different
in many ways from symbolic AI (and this can be seen in the errors and limitations of reasoning
in humans, as well as in our ability to learn to do this at scale, with distributed representations
and efficient search).} Yoshua Bengio, 2021.

\end{quotation}
}

A number of recent  breakthroughs  have shown  the remarkable
power of deep learning and deep reinforcement learning \cite{lecun2015deep,dl:cacm}.
These developments, however, have been   tied    to specific tasks like Chess, Go, or 
Atari games  \cite{dqn,silver2,silver2017mastering}. Progress in out-of-distribution generalization or in the generation of modular components that can
be assembled dynamically for different tasks,  has been more limited \cite{josh,darwiche,marcus1,geffner:ijcai2018}. 

While it is   assumed that  these  limitations   can  be overcome   by adding
suitable inductive biases in current neural network  architectures \cite{shanahan:review,bengio:high-level},
the notion of inductive biases itself  is often left vague and does not always provide meaningful guidance.
Traditionally, inductive biases refer to biases  in the hypothesis space, and in the case of neural networks, to
the structure of the parametric function  captured by the architecture. 
More recently, the notion  has been grounded  on the invariant properties of such functions \cite{invariant-nets}, 
but more often they are used  to refer to
intuitions that  are not spelled out in formal detail and  are not explicitly evaluated.

\Omit{
 For example, convolution nets capture group invariance over space,
graph neural nets capture equivariance of entities and relations, and so on.
In principle, one should use or device then new architectures in terms of the
invariants to be enforced.
}

In this paper, I aim to articulate a more abstract approach to representation learning 
where  the learned  representations are not those that emerge after training
a neural network, but those that result over  a given \emph{target representation language with a well understood semantics.}
The approach 
is  implicit in mainstream, symbolic AI, from  McCarthy's   observations  about the representation of  general  abstractions 
 \cite{mccarthy:1960}, to   Pearl's  emphasis on the  language required for reasoning about   causality \cite{pearl:bow}.
The challenge   is  to learn from data the representations  that have traditionally been crafted
by hand without having to appeal to background knowledge. 

The goals of the paper are  to make the  ideas behind the language-based approach to representation learning  explicit,
to place them in a broader context  where the design  of the target language is critical, and to illustrate them in the setting  of learning to act   and plan.
For this, after a general discussion, I   consider the problems of learning  actions,   general policies, and problem decompositions
over suitable domain-independent languages.

\Omit{
The rest of the paper  is organized in the following parts: 
1)~languages, semantics, and generality,  2)~an  example, 3)~causal models, 4)~languages,  models, and solvers,
5)~languages vs. inductive biases, 6)~related work, and 7)~a technical part from results mostly published elsewhere about 
learning general action models,   policies, and decompositions.
}

\Omit{
  The limitations of  ``symbolic AI''  are indeed   not in the
  representation languages per se but in their use. The challenge is to learn from data the representations  that have been
  written by hand, and  in some cases, coming up with new languages.
}

\section{Languages, Semantics, Generality}

\begin{quotation}
\noindent
\emph{It is hard to find a needle in a haystack, but it helps to know what a needle looks like.} J. Pearl\footnote{Pearl's  quotes
are from his twitter feed unless otherwise noted; \url{http://web.cs.ucla.edu/~kaoru/jp-tweets}.}

\end{quotation}

In  \emph{Generality in AI},  John McCarthy \citedate{mccarthy:generality},  one of the founders of the field,
quotes an early paper that says   that ``If one wants a machine to be able to discover an abstraction, it seems
most likely that the machine must be able to represent this abstraction in some relatively simple way''  \cite{mccarthy:1960}.
From this and the need to  use the learned abstractions in a \emph{flexible way}, McCarthy   concludes that
the representations have to  be expressed in a  logical language. 

Sixty years later, Yoshua Bengio, a  leader in  the field of deep learning interested in bridging the gap
between deep learning and high-level reasoning, addresses similar issues but in  slightly
different terms:\footnote{From  the abstract at \url{https://ijcai-21.org/invited-talks}.}
``Systematic generalization is hypothesized to arise from an efficient factorization of knowledge
into recomposable pieces corresponding to reusable factors \ldots This is related yet different
in many ways from symbolic AI'' \cite{bengio:high-level}.

Bengio's  point  that the research agenda that he describes for capturing high-level reasoning
is ``related yet different'' than  McCarthy's (symbolic AI) is certainly correct.
The claim in this paper, however, is that  there is much to be gained by making the two research
agendas  \emph{complementary}. In other words,  symbolic AI   has  developed families of formal languages
for ``factorizing knowledge into reusable  pieces'' with the right semantics for supporting composition and generalization. 
The limitation of symbolic AI is not in the languages themselves  but in their use by humans,
which as Bengio says, does not scale.  The challenge and the opportunity is to learn the  representations
over such languages (i.e., symbolic representations) directly from data.

Pearl's quote at the beginning of the section  refers  to  representations for causal reasoning.
I  aim to illustrate that his  point   is more general and applies to all representations that must be
learned and {combined}. The ``needles'' are the representations sought, and we know how they look like when they
are representations over known languages. Interestingly,   Bengio also  finds  language relevant for 
capturing the abstractions that are  required for combinatorial   generalization, but ``language''
for him, as for  others,   is natural language, not a formal language  with a  semantics. 

\section{Example}

\begin{quotation}
\noindent   \emph{Toy problems is where you learn if  you are on the right track. Non-toy problems
 is when you hide  you don't know  which track you are on}. J. Pearl
\end{quotation}

\begin{figure}
\begin{center}
   \includegraphics[scale=0.2]{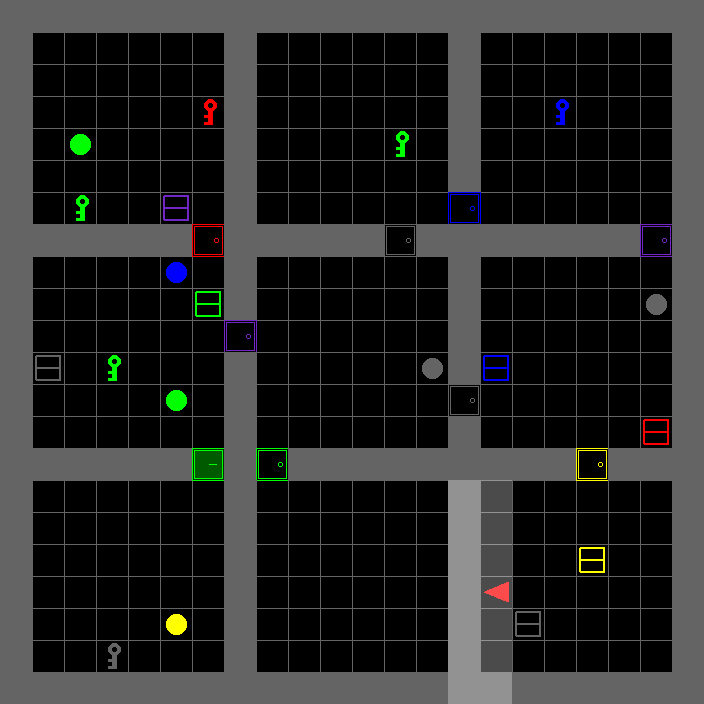}
\end{center}
\caption{\small Minigrid environment and a possible goal: \emph{pick up the grey box behind you, then go to the grey key and open  door.}
 The agent is the red triangle;  the light-grey area is his  field of view. Problem is similar to a ``classical planning problem''
 \emph{except}  that the domain predicates and action schemas are not known. In addition, goals are to be achieved reactively
 from learned general policy, not by planning \cite{babyAI}.}
\label{fig:minigrid}
\end{figure}

A simple  toy problem  will help us  to make the discussion of structural generalization  concrete. 
Figure~\ref{fig:minigrid} shows the Minigrid environment; a benchmark introduced for  learning
to interpret and achieve goal instructions  \cite{babyAI}.
As shown in the figure, a goal  may be: ``pick up the  grey box behind you, then go to grey key and open
the door''.  The agent is the red triangle  and the limited field of view is displayed  in light-grey.
The general  problem is to learn a  \emph{controller}  for the agent that accepts \emph{goals}  and  \emph{observations}, 
and outputs the \emph{action} to do in each  step for reaching the goals.

The Minigrid environment is similar to a classical planning problem  \cite{geffner:book,ghallab:book} except that 
the {action model}  and the {goal language} are  {not given}.
Both {supervised} and {unsupervised}  approaches have been  tried, 
and success has been  partial \cite{babyAI,minigrid}:
{millions of trials} are required to achieve the given goals,
and even then  {success rates} are not 100\%.
For improving  performance, intuitions leading to alternative
architectures and loss functions are introduced
(e.g., presence of objects or sparse interactions)
which are then  evaluated experimentally in relation to  baselines.
\Omit{
The key challenge for deep reinforcement
learning approaches  is to improve   success rates and  data efficiency, a   challenge that  is normally
addressed is by finding more suitable neural network architectures
and loss functions. For example, one can assume that a small number of objects  must
be tracked and that  the interactions among them  are  sparse \cite{rim}. The intuitions, however,
are  presented informally to motivate an   architecture that is  evaluated experimentally in
relation to suitable  baselines.
}
From a methodological point of view, this is not entirely satisfying, 
and two  key questions are  \emph{what} is it that we are trying to learn, and how
this object can be  characterized  mathematically, independently of its computation.
A   step in this direction is to notice  that we  look   for  a mapping from \emph{goals} $G$
into general policies  $\pi_G$ for achieving them  in a broad range of situations involving 
any number of objects at any locations, and hence, different state spaces. 
 
Even  this  toy example  is a hard problem  given  how little is known a priori. The surprise is not that supervised and unsupervised DL approaches struggle
in the problem but that they manage to  generate a meaningful behavior at all.
The immediate goal  for us is not to do better  in the task  but to identify the  building blocks
that are needed for  approaching this and other  problems  in a meaningful way.
Key questions  from the perspective of language-based
representation learning are what is a good, domain-independent
language for expressing the \emph{general policies} $\pi_G$, and
how can  representation in  that language be learned?
Related questions are what is a good, domain-independent  language for expressing the
\emph{dynamics} of  Minigrid  and how can representations over such a language  be learned?
The answers to these questions,  to be sketched  below,
do not have to get in the way of  developing  neural architectures
for solving these  problems; the hope  is that the answers
can inform our understanding  of  the  problems and their solutions,
neural and  otherwise.

\section{Causal Models}

\begin{quotation}
\noindent  \emph{
My greatest challenge was to break away from probabilistic thinking and accept, first, that people are not probability thinkers but cause-effect thinkers and, second, that causal thinking cannot be captured in the language of probability; it requires a formal language of its own.} J. Pearl\footnote{
\url{http://www.cambridgeblog.org/2012/07/qa-with-judea-pearl-part-one}.}
\end{quotation}

\Omit{
\begin{quotation}
 \emph{The equations for the spring example are typical: after Nature
spends some time, may be a billionth of a second, looking at $X$, multiplying it
by a constant, adding to it some noise, and deciding that $Y$ deserves the value $y$
(great work, mother Nature!), our job is to decipher the strategy of Nature.} Judea Pearl.\footnote{Turing Award lecture}.
\end{quotation}
}

One formal  language that is  making it  into mainstream ML is the language of (structural) causal models (SCMs). 
This owes to  the work of Judea Pearl and others that  has  revolutionized our understanding of causality by articulating
a simple  formal language  and a semantics to talk about causes and effects \cite{pearl:causality,pearl:bow}. 
The language accommodates observations, interventions, and counterfactuals. A \emph{structural causal model (SCM)} 
can be understood as a deterministic Bayesian network (conditional probabilities are all one or zero)
that defines not just one joint probability distribution over the variables  but  many.
A SCM   handles   interventions   (actions) of the form $do(X\!=\!x)$,  by which a   variable  $X$ is set to a  specific value $x$,
and the probability distribution that results from such   actions is the distribution that is encoded by the
``mutilated'' Bayesian network where the parents of variable $X$  are replaced by the single parent $do(X\!=\!x)$
for which  $P(X\!=\!x|do(X\!=\!x))=1$. The answers to queries about combinations of observations, interventions, and counterfactuals are  determined
by a SCM once the priors on  ``exogenous'' variables (those without parents) are given. The language of SCMs
has been used, for example,  to determine the conditions under which the answer to queries in one
causal model ``generalize'' or  ``transport''  to another causal model \cite{elias:transport}.

In principle, structural causal models can be learned from data, and this is a very active line of research, both in the
cases where the variables in the model are given, and when they are not   \cite{bernhard:causal}. 
This  does not mean, however,  that structural causal models can be  bypassed  when answering  queries  from  data.
In order to do that meaningfully, the design of the  algorithms must take the  semantics of SCMs into account   \cite{pearl:empirical-ml}.  
Doing this, model-free, while  ignoring the language and semantics of SCMs runs  the risk of reinventing the wheel  with not  much guidance,
as experimental evaluations are no substitutes for  a meaningful theory.
This does not rule out the possibility of learning causal  models using neural nets  by stochastic gradient descent,
but the  architecture and loss functions must be aligned with the representations sought. 

\Omit{ quote
There is an old saying in AI   "you cannot learn what you can't represent", usually credited to Gerry Sussmann,
which can be understood as saying that learned representations must be learned over a language, if they are to be
used in a flexible manner. The main advocate of this line of thinking in recent times has been Judea 
}

\section{Languages, Models, and Solvers}

\begin{quotation}
\noindent  \emph{Every science that has thriven, has thriven upon its own symbols.} De Morgan (1864), quoted by J. Pearl.\footnote{
\cite{pearl:bow}.}
\end{quotation}

Bayesian networks and structural causal models are models that make predictions 
from knowledge expressed in terms of  variables, graphs,  and  probabilities.
In AI, other languages and models have been developed as well
some of which are relevant to our focus on actions  and planning.
\Omit{
There is indeed  a common misconception about AI  which assumes that
it  is all learning or knowledge-based. This misses an  important  thread
which has  centered on   the formulation of general mathematical models and
their corresponding solvers. These  models include Bayesian networks and structural causal models,
but  also constraint satisfaction problems (CSPs), SAT and answer set programs, classical planning, and fully and partially observable MDPs, among others.
In these model-based approaches, knowledge is   about the  model, which  is defined in compact form through
 suitable  languages  \cite{geffner:solvers,geffner:ijcai2018}.
}
For example, classical planning refers to  planning with deterministic actions with known effects and preconditions, 
from a known initial state,  given a compact  encoding of the actions in terms of state variables.
These encodings have a size that is polynomial in the number of variables
but   result  in state models of exponential size. Compact languages for representing other state
models like MDPs and POMDPs, have also been developed.

The use of  languages for encoding state models has been motivated by two  reasons.
First, state models need to be specified in a concise manner, as they would not fit
in memory otherwise. Second, it is assumed that a compact specification  reveals structure
that can be exploited computationally. For example, a common technique for solving
classical planning problems is using  heuristic functions,
yet these heuristics can be extracted from  compact representations
but  not  from flat models \cite{mcdermott:unpop,bonet:aij-hsp}.

The benefits of languages supporting  compact  action representations, however, 
go well beyond the facilities that they provide for model specification and computation,
as they also provide the  ingredients needed for \emph{generalization, transfer,
and knowledge reuse.} Indeed, these languages have been designed   for human to use  with these goals in mind:
when writing the description of a planning problem, we want the description to be reusable with
minor modification in  similar problems.
The use of \emph{first-order languages} for referring to  objects and relations
has been essential for this purpose. 

Consider for example, a simplification of the Minigrid domain, that we refer
to as Delivery, where there are $N$ objects in a grid  $n \times m$, and the
goal  is to pick up the objects, one by one, and  deliver them to  a target cell.
The actions available are to pick up and drop an object, and to move one cell at a time.
Different instances   of the Delivery domain are  encoded in planning languages such as
PDDL in two parts as $P=\tup{D,I}$. One part, the domain $D$, encodes what is common about
all the Delivery  instances in terms of three  \emph{action schemas}: 
\emph{pick}, \emph{drop}, and \emph{move}. The other part,  $I$, details  the objects in the instance, the ground atoms
that are initially true, and those that must be made true in the goal \cite{mcdermott:pddl,pddl:book}.
For example, the action schema \emph{pick} can be defined as:

\begin{tabular}{l}
\ \\
$pick(o,x,y)$: \\
\ \ \ \emph{Prec:} $at(o,x,y), at_r(x,y)$, $handempty$ \\
\ \ \ \emph{Eff:}  $hold(o)$, $\neg at(o,x,y)$, $\neg handempty$ \\
\ \\
\end{tabular}

\noindent where $o$, $x$, and $y$ are the schema arguments,
and  preconditions and effects are  atoms  formed by predicate symbols
and some of the schema arguments. The schema for $move$
can be expressed in turn as:

\begin{tabular}{l}
\ \\
$move(x,y,x',y')$:\\
\ \ \ \emph{Prec:} $at_r(x,y)$, $adjacent(x,y,x',y')$ \\
\ \ \ \emph{Eff:}  $at_r(x',y')$, $\neg at_r(x,y)$\\
\ \\
\end{tabular}

\noindent where $at_r$ and $at$ are the predicates that encode the location of the agent and
the packages, and  $adjacent$  encodes the grid topology. 
\Omit{
The Delivery \emph{domain}  $D$ would be  given by  the  schemas
for the actions $pick$, $drop$, $move$, and a Delivery \emph{instance} $P=\tup{D,I}$,
by the domain $D$ and  information $I$ detailing the objects, the ground atoms
that are initially true, and those that must be made true in the goal.
}
These $pick$ and $move$   action schemas, along with the $drop$ scheme  and the predicates involved in them,
are  precisely  the  ``reusable pieces''  over  all Delivery  instances, and hence  if
we want to learn a dynamic model from some instances that   generalizes   to other instances,
we will be well advised to learn  a representation of this type.

There are indeed  important similarities between  planning  languages  and  structural causal models:
SCMs provide  a \emph{compact and invariant  description} of the effects of interventions on probability distributions,
while  planning languages provide a \emph{compact and invariant description} of the effects of interventions on states.
Compact, first-order  languages for defining  probabilistic graphical models,  MDPs,  and POMDPs have also been developed
\cite{kersting1,kersting2,ppddl,rddl}, and if we want to learn \emph{models that generalize}, such  languages would
be good targets for learning as well.


\section{Languages vs. Inductive Biases}

\begin{quotation}
\noindent  \emph{It's hard to understand why we should struggle to understand deep learning instead of learning
deep understanding}. J. Pearl
\end{quotation}

There is a compelling reason  for why learning approaches are either  model-free and learn no models,
or are model-based but do not learn  language-based  representations (i.e., do not learn symbolic
representations). The reason is that learning such representations appears  to require humans
in the  loop, something that gets in the way of  the automated learning  pipeline. This impression, however, is wrong:
while   \emph{languages} like those of   SCMs and planning have been developed to be used by humans, this  does not
mean that the \emph{representations} over such languages can   only be provided by humans  and cannot
be learned  from data. Certainly, there are obstacles to  overcome for achieving
this and a key one is the identification of the \emph{(state) variables} from  unstructured data,
but this is a technical problem that can be solved. I'll show  that there are indeed crisp solutions to this problem
that exploit the  natural inductive biases  of language-based representations. 


Current DRL approaches can learn in principle policies that solve problems such
as Delivery or Minigrid for any value of the parameters, but even then,
it is not clear why this is so. For representations learned over languages
designed to support modularity and reuse, the answers to these  questions
follow from their  semantics.

While the use of formal languages for learning representations is not common
in deep learning, there is an increasing  trend to reflect  intuitions about 
the representations  sought in the architectures. For example, RIM networks
assume a dynamics determined by sparse object interactions \cite{rim}.
Yet, informal talk of sparse interactions is no substitute for a language with a clear
semantics that can represent the range of possible sparse interactions and
lead to  representations that can be understood in that way.

As mentioned before,  suitable target languages   yield   meaningful learning biases, as generalization  is
most often the result of learning compact descriptions. In Bayesian networks,  compactness comes from sparse graphs,
while  in SCMs, compactness comes from the  language and  semantics of interventions.
In planning,  compact descriptions  result from the language of action schemas and the  predicates used in them.
Compact  descriptions are  easier to learn  and afford a   powerful generalization, implying
that  representation size is a key  bias in language-based representations learning.
While there is no similar notion  in deep  learning, the  ``right''  biases  in neural nets
would be the ones  that deliver   compact and reusable representations of this type.

\section{Related Research}

Language-based representations, most often first order,  are at the heart of the models and solvers
studied in AI \cite{geffner:solvers}. 
The problem of learning such representations
from data is active in some of these settings, like learning causal representations, mentioned above, and
learning general action models, to be discussed below. Methods designed  to learn symbolic  representations from data
provide  a  natural way for integrating learning and reasoning \cite{konidaris:jair,making-sense}.
Neuro-symbolic  methods   make use of prior symbolic knowledge \cite{luciano:ltn,asp:nnet},
possibly in terms of first-order probabilistic models \cite{kersting1,luc:neuro}. 
The use of  logical languages has  been central  in recent  characterizations of the expressive power of graph neural networks
\cite{barcelo:gnn,grohe:gnn}. Domain specific task languages   have  been used 
in a number of settings \cite{josh,josh:logical,josh:gvgai},
including the dynamics of  MDPs \cite{oo-mdp}. These  languages are  domain specific in the sense
that they  assume a particular vocabulary. 
Language-based representations of rewards have been explored too  \cite{sheila:rewards,giuseppe:rewards}.
Some deep learning approaches aim to   approximate   first-order  formulas
like conjunctions of atoms  \cite{shanahan:predinet} or  rules \cite{bengio:rules},
while  others draw intuitions from them \cite{shanahan:review,bengio:high-level}.
Generally, symbolic methods like those considered in inductive logic programming
\cite{ilp} assume and exploit background knowledge, while deep learning approches
do not use and  do not produce  background knowledge; i.e., knowledge that
can be reused. This is  their   advantage and also their  limitation.

\Omit{
\item \textbf{Causality:}  Pearl, Bernhard ..: causality, and more generally,  importance  of model-based and importance 
of languages .. Invariant Risk. Peters
\item \textbf{Learning and languages:} all those below, and STAR/learning lifted representations, etc. (Kersting, Poole, ..)
Shanahan predinet. 
\item \textbf{Domain specific languages:} Domain SL Josh: including Atari Pedro ..
\item \textbf{Learning first-order models:} Also: exploit languages for expressing rewards, solvers, factored, OO-MDPs .. Konidaris. Matusaro.
  Giuseppe. Sheila. Languages for Rewards. 
** Konidaris, Asai, Grohe, Simon, Shanahan, Kersting, (previous list), .. Meta-learning: quickly adjust pre-trained models ..
}

\section{Action models, policies, and decompositions}

We consider next three   concrete examples of domain-independent languages for acting and planning, and how representation over them  can be learned.
\Omit
{We  consider  general action models,   general policies, and general  problem decompositions, none tied to specific domains. 
The language  for actions is  well known but not those for  policies and decompositions, illustrating that languages for many settings
cannot be taken off the shelf and must be designed. It is important to distinguish the languages themselves from representations
in the language, learned or hand-crafted. The first are tied to a class of models and are domain-independent; the second, are
tied to a class of models and a particular domain.
}


\subsection{Languages for general action models}

Classical planning problems $P=\tup{D,I}$ are described in terms of 
a planning domain $D$  involving  action schemas and  predicates, 
and instance information $I$ detailing the objects, the initial situation, and the goal.
\Omit{
This is a representation
that is particularly convenient for learning  general policies, as  all the 
instances are assumed to come all from the same domain (same action schemas and predicates).
Some of the schemas for the Delivery domain were illustrated above; particular
instances are  obtained by selecting a number of packages and a  grid,
placing the packages and the agent in some location, and defining their goal locations.
This is a simple language for planning with deterministic actions with discrete effects (times and values),
which corresponds to lifted STRIPS with negation, a fragment of the PDDL standard; richer languages could be used as well.
}
A planning instance $P$ defines  a unique graph $G(P)$ where the nodes $s$  stand for the
states over $P$ and edges  $(s,s')$ express   that there is a ground  action $a$ in $P$
that maps the state $s$  into $s'$. The states $s$ are the  possible truth valuations over the ground atoms in $P$.
The general \emph{model learning problem over this language}  can  be expressed then  as the
following inverse problem \cite{bonet:ecai2020}:



\vspace{.1cm}

\noindent \framebox{\begin{minipage}{.95\linewidth}
\noindent Given plain graphs $G_1, \ldots, G_k$,  find the simplest domain $D$ and instances $P_i=\tup{D,I_i}$
over $D$ such that the given graphs  $G_i$ and $G(P_i)$ are isomorphic, $i=1, \ldots, k$.
\end{minipage}}

\

Variations of this problem have  also been  considered where the edges in the input graphs $G_i$ are labeled
with  action types   (e.g., $pick$, $drop$, $move$), and  edges may be missing or  observations may be
noisy \cite{ivan:kr2021}. The  domains   learned from some instances
can then be used to predict the effects of actions in other, unseen instances.
This  learning  formulation has been used to obtain the predicates and action schemas
for domains such  Blocks, Tower of Hanoi, and others.
For example, the following domain description for Hanoi is learned from a single graph, produced by an  instance
with  3 disks (predicate and variable names are ours):

\noindent \begin{tabular}{l}
  \ \\
 \textit{Move}$(d,fr,to)$ \\
 \  Sta: $neq(d,to)$, $neq(d,fr)$, $neq(to,fr)$, $\neg larger(to,d)$\\
 \  Pre: $\neg p(to,d)$, $\neg p(fr,fr)$, $p(d,d)$, $p(to,to)$, $p(fr,d)$ \\
 \  Eff: $\neg p(to,to)$, $\neg p(fr,d)$, $p(t,do)$, $p(fr,fr)$ \ . \\
 \ \\
\end{tabular}

The domain learned can be shown to be  correct for instances of any size (any number of disks and pegs),
and use the predicate $p(x_1,x_2)$ for two different purposes: for capturing the relation
$on(x_2,x_1)$ when $x_2 \not= x_1$, and for capturing  clear($x_1$) otherwise.
The three schema arguments $d$, $fr$, $to$ represent disks, and ``Sta(tic)''
refers to precondition atoms that are not affected by the actions.
One can actually  test that this domain description $D$  is correct
experimentally, using (validation)  graphs $G$ obtained from other instances, and checking
if there is an $I$ such that  $G$ and $G(P)$ for $P=\tup{D,I}$ are isomorphic
(a simpler version of the learning  problem above). The learned representation
also identifies  the \emph{state variables} of the problem  through the
$p(d,d')$ atoms that encode the location of each of the disks $d$.
As  discussed by  \citeay{bonet:ecai2020},  the use of a first-order target language
with action schemas is critical for learning such state variables, as propositional
representations cannot be reused in the same way and hence do not admit the same
learning  bias.

The  language of  action models (action schemas and predicates) is  suitable for learning
in this setting,  not just because it supports representations that  generalize  to other instances,
but also  because it  defines   a \emph{heavily biased hypothesis space}, with the space of possible domains
being characterized by {a small number of parameters}  with small integer values, like 
the  number of action schemas and predicates and their arities. Provided with 
bounds on these values, the learning problem becomes a combinatorial optimization problem that can be solved in a
number of ways, in many cases optimally. The  optimization criterion  used by \citeay{ivan:kr2021}, for example,
minimizes   the sum of actions  and  arities. \citeay{asai:fol}, on the other hand, learn first-order action representations
using deep learning, while  \citeay{locm}   learn  first-order representations  heuristically
assuming that action arguments in state transitions are observable. Many other works learn similar
representations but assuming that the domain predicates are known.

\Omit{
The hope is that, eventually, the move from simple, combinatorial
optimization approaches to  deep learning approaches that   learn the same type of structured  representations, can be done in a
principled manner,  for delivering learning methods that are robust and scalable, and that result in representations that are crisp and transparent,
and which generalize due to their structure and semantics. 
}

\subsection{Languages for general policies}

The target languages for learning in many settings  can be taken off the shelf like the languages for
representing actions and causal models discussed above. But for other  tasks, new domain-independent languages
with the right properties may have to be created. For example,  in  the Minigrid problem, DRL approaches are not after
general dynamic  models, but after general policies:  policies that can deal with \emph{any} instance of the domain. 
What is then  a good language for representing such policies that is not tied to this  particular domain?
This question has been considered in the area of  \emph{generalized planning}  \cite{srivastava08learning,hu:generalized},
and the language below follows the one introduced by \citeay{bonet:ijcai2018}.

A general policy $\pi$ for a class of domain instances  $\Q$ is 
a set of \emph{policy rules} of the form $C \mapsto E$ where $C$
contains boolean conditions of the form $p$, $\neg p$, $n=0$, or  $n > 0$,
and $E$ contains effects of the form $p$, $\neg p$, $p?$, $n\mminus$, $n\pplus$, $n?$,
where $p$ and $n$ stand for boolean and numerical \emph{features}.
Features are  functions over states. 
Boolean features $p$  can have value \emph{true} or \emph{false}, 
and  numerical features  $n$ can have  any non-negative integer value.
Conditions in $C$ like $p$ and $n=0$ are true in a state when $p$ has value true, 
and $n$ has value $0$ respectively, and effects in $E$ like
$p$ ($\neg p$), $n\mminus$ ($n\pplus$), and $p?$ ($n?)$
indicate that $p$ must be made true (resp. false), that $n$ must decrease (resp. increase), and
that $p$ (resp. $n$) can change in any way. Features not appearing in the effects of a rule
must  keep their values. The value of all the features  $\Phi$ in a state $s$ is expressed as $f(s)$, 
and $f$ without a state argument refers to an arbitrary feature valuation.

A \emph{pair of feature valuations} $(f,f')$ satisfies a policy rule $C \mapsto E$
if $f$ makes the conditions in $C$ true, and the change in feature values from  $f$ to $f'$
is compatible with $E$. 
A  \emph{state transition} $(s,s')$ in  $P$ is  compatible with a policy  $\pi$ if $(f(s),f(s'))$ satisfies a policy rule,
and a \emph{state trajectory} $s_0, \ldots, s_n$ is compatible with the policy in $P$  if $s_0$ is the initial state of $P$
and each transition $(s_i,s_{i+1})$ is compatible with $\pi$. 
Finally, the  policy $\pi$  \emph{solves}  $P$  if every maximal state trajectory compatible with $\pi$ reaches a goal state of $P$,
and it solves $\Q$ if it solves every instance $P$ in $\Q$.

For example, the policy $\pi$ over  the  features $\Phi=\{H,n\}$, captured by the following two rules

\begin{tabular}{ll}
\ \\
$\prule{\neg H, \GT{n}}{H, \DEC{n}}$;  & pick block above $x$ \\[.1cm]
$\prule{H, \GT{n}}{\neg H}$;  & put block away\\
\
\ \\
\end{tabular}

\noindent where $H$ is true if holding a block and $n$ is the number of blocks above a block $x$,
solves  the  class $\Q$ of Blocksworld problems where the  goal is $clear(x)$, 
regardless of the number or initial configuration of blocks.
The first rule in the policy says to do any action that makes $H$ true and decreases the value of $n$, provided
that $H$ is false and $n$ is positive, while the second rule
says to do any action that makes $H$ false and does not affect the value of $n$, 
provided that $H$ is true and $n$ is positive. 

More interestingly, a general policy for the Delivery domain above
can be defined using the features   $\Phi=\{H,p,t,n\}$ for  ``holding'',  ``distances to nearest package and target'',
and ``number of undelivered packages'', as:

\begin{center}
\begin{tabular}{ll}
\ \\
   $\prule{\neg H,\GT{p}}{\DEC{p},\UNK{t}}$;  & go to nearest pkg \\[.1cm]
   $\prule{\neg H, \EQ{p}}{H}$;  & pick it up \\[.1cm]
 $\prule{H,\GT{t}}{\DEC{t}}$;  & go to target \\[.1cm]
 $\prule{H,\GT{n},\EQ{t}}{\neg H, \DEC{n}, \UNK{p}}$;  & drop pkg.  \\[.1cm]
\end{tabular}
\end{center}

The first rule says  to do any action that decreases the distance $p$ to the nearest package  when not holding
a package and the distance is positive, whatever the effect on the distance $t$  to the target. The reading of the other rules is similar.
These are policies written by hand though, and the question is how  such   policies can  be learned?
As before, this learning  problem has been formulated and solved as a combinatorial optimization problem by creating a  large but finite
set $\cal F$ of possible boolean and numerical features  \emph{from the  domain predicates},
using a standard grammar based on description logics, which is a fragment of 2-variable logics \cite{description-logics}.
Provided with this set  $\cal F$ where each  feature  is given a cost
(the number of grammar rules used to derive it),  the learning task becomes: 

\

\noindent \framebox{\begin{minipage}{0.95\linewidth}
\noindent Given a domain $D$, instances $P_1, \ldots, P_k$ of $\Q$, and a finite pool of features ${\cal F}$,  each
with a cost, find the cheapest set of features $\Phi \subset {\cal F}$ and a policy $\pi$ over them 
such that $\pi$ solves the instances $P_1$,  \ldots, and $P_k$.
\end{minipage}}

\

This is a  combinatorial optimization problem that is cast and solved as a Weighted Max-SAT task
\cite{frances:aaai2021}.
Once again, the language in which representations are sought provides a \emph{strongly biased
hypothesis space}  where policies that involve few simple features (in terms of the domain
predicates) are  preferred.
As before, nothing precludes the use of deep learning   to provide an alternative
computational method, potentially more scalable and robust \cite{sylvie:asnet,mausam:dl2}.
A formal  step in this direction  is the computation of general optimal value functions
using graph neural networks \cite{simon:gnns}, that exploits a correspondence between
2-variable logics and GNNs \cite{barcelo:gnn,grohe:gnn}.

\subsection{Language for decomposing goals into subgoals}

\Omit{
A  general policy tells the agent what action to do in every  (reachable) state $s$ over
a  class of problem $\Q$; namely, it says to  do any action that maps $s$ into a state $s'$
such that the resulting pair of feature valuations $(f(s),f(s'))$ satisfies a
policy rule $C \mapsto E$. In many cases, however, computing such policies may be  too hard or impossible.
A common alternative  then to general  policies, that solve each  instance  reactively,
and planning, that solves each instance by search, is a general  strategy for
decomposing    goals  into \emph{subgoals} which  can then be achieved by a
\emph{bounded}  (polynomial as opposed to exponential) search.
In this way rather than transferring a general policy from instance to instance in $\Q$, one   transfers
the \emph{common subgoal structure}.
}

The problem of \emph{expressing} and \emph{using} the common subgoal structure of a collection of problems $\Q$
has been important in AI since the 1960s   \cite{simon:gps,htns}, while the problem of \emph{learning} such
 structure (in the form of  intrinsic rewards) has become  important  in recent  RL research as well   \cite{singh:rewards}.
We are interested in a similar  problem but want to learn such structure over a suitable formal language.

A \emph{policy sketch} or simply a \emph{sketch} is a set of  sketch rules  $C \mapsto E$
of the same form as   {policy rules}. But while policy rules   filter  1-step transitions; namely,
when in a state $s$,  select a 1-step transition to any $s'$ such that the feature valuations $(f(s),f(s'))$ satisfy a policy rule,
sketch rules define \emph{subproblems:} when in a state $s$, reach a state $s'$, \emph{not necessarily in one step},
such that the feature valuations $(f(s),f(s'))$ satisfy a sketch rule  \cite{bonet:aaai2021}.

Sketches  \emph{decompose} problems  into \emph{subproblems} without prescribing how these
subproblems should be solved (going from $s$ to  $s'$).  One is interested, however,  in sketches
that yield  subproblems that can be solved efficiently,  in low polynomial time
(in the number of problem variables), and this is  guaranteed
when   subproblems  have  a   low,  bounded width \cite{nir:ecai2012}. 
In that case, the sketch has a bounded width, and all the problems in $\Q$
can be solved in polynomial time. 

For example, a sketch $R_1$ for Delivery that involves the single feature $n$
which  tracks the number of packages not yet delivered, is given by the rule

\begin{center}
\begin{tabular}{ll}
\ \\
$R_1: \{\prule{n > 0}{n\mminus}\}$
\end{tabular}
\end{center}

\noindent that expresses a decomposition where, in states $s$ where $n > 0$,  states $s'$ should be reached
where the value of $n$ is lower than in $s$.  One can show that the resulting  subproblems have
a width bounded by $2$.  Likewise, a sketch $R_2$ over the  features $n$ and $H$ with the same
meaning as above, can be given with two rules:

\begin{center}
\begin{tabular}{ll}
\ \\
$R_2: \{ \prule{\neg H}{H} \, , \, \prule{\GT{n},H}{n\mminus,\neg H}\}$ \ . 
\end{tabular}
\end{center}

\noindent  The rule on the left says  that  if  not holding a package, get hold of one, 
while the other rule, that  if holding a package,  deliver it. The sketch $R_2$
has width $1$ meaning that all subproblems and hence all  Delivery instances
are rendered   solvable  in linear time.
\Omit{
that it  decomposes the problem into subproblems of width $1$. There is a general planning
algorithm that  solves  the problems $P$  in $\Q$ in  time
$O(bN^{|\Phi|+2k-1})$, in the worst case, provided a sketch of  width $k$ , 
where  $\Phi$ is the set of features,   $b$ is the average number of actions per state,
and $N$ is the number of atoms in $P$. For $k=1$, this means $O(bN^{|\Phi|})$, where $\Phi$
is a constant that does not depend on the size of  $P$.
}
The uses of hand-crafted sketches  and
the learning of sketches have been addressed
by  \citeauthor{dominik:kr2021} \citedate{dominik:kr2021,dominik:learning-sketches}.

\section{Summary}

Deep learning and deep reinforcement learning are incredibly powerful techniques
that struggle with structural generalization. While  researchers assume that the right
inductive bias in the architecture is all that is needed, no much guidance is offered
to get there. In this paper, I've argued that learning  representations over suitably designed
formal languages with a semantics provides  a research path that is crisp and meaningful,
and illustrated the approach by focusing on the problems of learning general action models, policies,
and subgoals (intrinsic rewards).  This is all  compatible  with   Bengio's vision that
systematic generalization arises  from an  ``efficient factorization of knowledge
into recomposable pieces'', but   complements  it by assuming that the pieces are expressed in a language. 
\Omit{
The challenge is to learn from data
the representations over these languages  that have  traditionally been crafted by hand.
This, at least, when  the  languages are taken off the shelf (SCMs, action schemas, first-order probabilistic models, etc).
In other cases, the languages  may have  to be invented  (general policies, sketches).
} 
Learning language-based representations  from data, indeed, is not incompatible
with the use of deep learning techniques.  Moreover, the integration of 
language-based representations and deep learning, one  describing what needs to be learned, and the other,
delivering it at scale, has  the  potential to inform the design of deep learning
methods  that  are more transparent  and which can be assessed in ways that go beyond performance curves. 

\Omit{

- Scalability, principle move from combinatorial optimization to deep learning: Simon

- New languages: limitations of main planning languages, GVG-AI, task and motion planning ..

- Learning grounded representations

. Among others. Reusable pieces, rol language, semantics ..

*** A challenge for the future is to use approximated, scalable
SGD approaches for learning crisp representations over the given target language. Thus we are just advocating to
encode prior knowledge via a target language with a clear and compelling semantics, both as a way to learn
meaningful representations that generalize and as a way to inform in a formal way the choice of neural network
architectures and loss functions.
}


\section*{Acknowledgments}

I thank Judea Pearl, and not just for the quotations.
I also  thank Blai Bonet  for comments and collaboration, the RLeap team,
and the reviewers. 
The research  is partially supported by an    ERC Advanced Grant  (No 885107),
the EU Horizon 2020 project TAILOR  (Grant No 952215), and 
the Knut and Alice Wallenberg (KAW) Foundation, under the WASP Program.

\bibliography{control}

\end{document}